\documentclass[lettersize,journal]{IEEEtran}
\usepackage{amsmath,amsfonts}
\usepackage{algorithmic}
\usepackage{algorithm}
\usepackage{array}
\usepackage[caption=false,font=normalsize,labelfont=sf,textfont=sf]{subfig}
\usepackage{textcomp}
\usepackage{stfloats}
\usepackage{url}
\usepackage{verbatim}
\usepackage{graphicx}
\usepackage{cite}
\usepackage{booktabs}
\usepackage{multirow}
\usepackage{multicol}
\usepackage{graphicx}
\usepackage{algorithm}
\usepackage{algorithmic}
\usepackage{wrapfig}

\begin{document}

\title{Context Gating in Spiking Neural Networks: Achieving Lifelong Learning through Integration of Local and Global Plasticity}

\author{Jiangrong Shen,
Wenyao Ni,
        Qi Xu,
        Gang Pan,
        and~Huajin Tang
\thanks{This work was supported by National Key Research and Development Program of China (No. 2022YFB4500100), National Natural
Science Foundation of China under Grant (No.
62306274, No.62236007), and Fellowship from the China Postdoctoral Science Foundation under Grant (No. 2023M733067 and No. 2023T160567). } 
\\\thanks{Jiangrong Shen, Wenyao Ni, Huajin Tang, and Gang Pan are with the State Key Lab of Brain Machine Intelligence, College of Computer Science and Technology, Zhejiang University, Zhejiang, 310027, China. }
\thanks{Qi Xu is with the Faculty of Electronic Information and Electrical Engineering, School of Artificial Intelligence, Dalian University of Technology, Dalian 116024, China.}
        }


\markboth{Journal of \LaTeX\ Class Files,~Vol.~14, No.~8, August~2021}%
{Shell \MakeLowercase{\textit{et al.}}: ALADE-SNN: Adaptive Logit Alignment in Dynamically Expandable Spiking Neural Networks for Class Incremental Learning}

\IEEEpubid{0000--0000/00\$00.00~\copyright~2021 IEEE}

\maketitle

\begin{abstract}
  Humans learn multiple tasks in succession with minimal mutual interference, through the context gating mechanism in the prefrontal cortex (PFC). The brain-inspired models of spiking neural networks (SNN) have drawn massive attention for their energy efficiency and biological plausibility. To overcome catastrophic forgetting when learning multiple tasks in sequence, current SNN models for lifelong learning focus on memory reserving or regularization-based modification, while lacking SNN to replicate human experimental behavior. Inspired by biological context-dependent gating mechanisms found in PFC, we propose SNN with context gating trained by the local plasticity rule (CG-SNN) for lifelong learning. The iterative training between global and local plasticity for task units is designed to strengthen the connections between task neurons and hidden neurons and preserve the multi-task relevant information. The experiments show that the proposed model is effective in maintaining the past learning experience and has better task-selectivity than other methods during lifelong learning. Our results provide new insights that the CG-SNN model can extend context gating with good scalability on different SNN architectures with different spike-firing mechanisms. Thus, our models have good potential for parallel implementation on neuromorphic hardware and model human's behavior. 
\end{abstract}

\section{Introduction}


The human brain can continuously encode and preserve new memories without disrupting previously acquired ones, an ability known as lifelong learning. 
This ability is believed to be closely related to the cognitive control mechanism in the primate prefrontal cortex (PFC). Cognitive control involves selecting context-appropriate tasks and executing them with minimal interference by gating task-irrelevant 
input dimensions \cite{miller2001integrative,flesch2022modelling}, thus ensures the brain dynamically allocates resources and creates specialized pathways for different tasks. 
Despite extensive research, it remains unclear how cognitive control enables that synaptic plasticity induced by past experiences of old tasks is maintained when new experiences of new tasks are acquired. 
Inspired by spike-based computation and plasticity learning in the brain, spiking neural networks (SNN) exhibit unique advantages in terms of energy efficiency, spatiotemporal dynamics, and biological plausibility computation \cite{maass1997networks,subbulakshmi2021biomimetic,kudithipudi2022biological,yin2023accurate,pei2019towards}. 
However, most of current SNN models are significantly affected by catastrophic forgetting, often requiring the random mixing of all task samples to learn new tasks effectively \cite{zhang2018highly,xu2021robust,fang2021incorporating,fang2021deep}. This approach, just like traditional artificial neural networks (ANN), is fundamentally different from human's multi-task lifelong learning manner. One possible reason is that these models update all weights simultaneously by only one plasticity rule, regardless of their relevance to different tasks. Therefore, combining the multi-level of local and global synaptic plasticity of SNN, we aim to explore solutions for implementing lifelong learning in SNN with context gating, and meanwhile attempt to model the cognitive control phenomenon of humans using SNN.



The most essential point in lifelong learning is the capability to perform competitively on the previous task after learning on the subsequently observed dataset. Vanilla SNN trained by global plasticity of backpropagation with surrogate gradients typically struggle to learn multiple tasks in the block learning manner (as depicted on the right side of Fig. \ref{fig_model_framework}). Because the network updates connection weights that were optimized for previously learned tasks to minimize the error for the new task, it leads to the loss of old knowledge. Therefore, the network can only preserve multiple tasks under the interleaved learning regime.
When encountering multi-task lifelong learning, those vanilla SNN often suffer from catastrophic forgetting, and their performance decreases sharply on previously encountered tasks after learning new ones. To overcome that problem, improved SNN for lifelong learning have been proposed, such as memory reserving \cite{mozafari2019bio}, regularization-based \cite{skatchkovsky2022bayesian}, reward-based \cite{allred2020controlled}, or neuromodulation-inspired synapse updating methods \cite{zhang2023brain}. 
Most existing models overlook the reconfiguration of input information across different tasks. Drawing inspiration from the cognitive control mechanism, which involves the PFC neurons selectively gating task-irrelevant information to execute context-appropriate tasks, we explore how to construct Spiking Neural Network (SNN) models with context gating for lifelong learning. Our approach integrates both local and global plasticity mechanisms, allowing the models to reconfigure the input information of different tasks continuously. Furthermore, we replicate experimental observations from human behavioral studies, to provide insights into the cognitive control mechanism through more biologically plausible SNN.




In this paper, our goal is to implement SNN with context gating for lifelong learning by integrating local and global plasticity,
and explore whether the proposed SNN could replicate the experimental observations documented in human participants. 
To this end, we develop single-spike and multi-spike SNN to show our model's applicability on different structures and spike-firing mechanisms. 
The contributions are as follows:







\begin{itemize}
    \item The framework of context-gated spiking neural networks (CG-SNN) is developed for lifelong learning by integrating local and global plasticity. CG-SNN alleviates catastrophic forgetting to a certain extent via local plasticity-induced context gating, instead of manual weight updating fixation or the samples having been learned.
    \item The single-spike and multi-spike CG-SNN models are implemented by integrating the global backpropagation plastivity with the local plasticity of spike-timing-dependent (STDP) and Oja learning rules, respectively. The iterative training between global and local plasticity is designed to strengthen the connections between task neurons and hidden neurons to preserve the multi-task relevant information. 
    \item The experiments are conducted to analyze the reasons for the effectiveness of context gating during SNN' lifelong learning. Besides, the performance of the proposed model is compared with the characteristics of human behavior in the cognitive experiments. The results suggest that the above two show a considerate degree of consistency. Moreover, the proposed model shows the prior biological plausibility of task neurons' selectivity in neural dynamics.

\end{itemize}

\section{Related Works}

Lifelong learning aims to remember previously trained old tasks and acquire new knowledge by training new tasks. The direct trained SNN models suffer from the catastrophic forgetting problem when facing multiple-task continuous training \cite{hao2023reducing, chen2023unified, hao2023bridging}. Different approaches have been developed to overcome catastrophic forgetting. 


The regularization method preserves synaptic connections that are deemed to be essential to consolidate previous knowledge. 
Considering synaptic noise and Langevin dynamics, the regularization-based method for SNN in \cite{ANTONOV2022512} determines the importance of synaptic weights by stochastic Langevin dynamics with local STDP instead of gradient. Based on Bayesian, each synaptic weight is represented by parameters that quantify current epistemic uncertainty resulting from prior knowledge and observed data. The proposed online rules update the distribution
parameters in a streaming fashion\cite{skatchkovsky2022bayesian}.
%


The memory replay method maintains representative samples or anchors with the key features of previous tasks as episodic memory to promote lifelong learning. SNN trained by R-STDP eliminate catastrophic forgetting with small episodic memory in  \cite{mozafari2019bio}. The cortical connections between excitatory neurons are plastic and regulated by STDP.
The results show that sleep can prevent catastrophic forgetting through spontaneous reactivation (replay) of both old and new memory traces.


Other lifelong learning methods of SNN are proposed based on the observed neuroscience mechanisms. It is 
The challenge of lifelong learning stability-plasticity dilemma challenge which means ensuring that the system can continue to
quickly and successfully learn from and adapt to its
current environment while simultaneously retaining and
applying essential knowledge from previous environ-
ments \cite{allred2020controlled,grossberg1987competitive}. Inspired by the dopamine signals in mammalian brains, the controlled forgetting network model is introduced to address the stability-plasticity dilemma with local STDP learning to control the forgetting process and reduce the accuracy degradation on new tasks \cite{allred2020controlled}.
Especially, context gating has been suggested to be effective in supporting lifelong learning. For instance, inspired by that neuron in PFC 
codes for specific tasks and exerts top-down control to prioritize context-appropriate stimuli and action, the context-dependent gating signal along with the input is introduced to guarantee that only sparse, mostly non-overlapping patterns of neurons are active for each task in \cite{masse2018alleviating}. In this method, between 80$\%$ and 86.7$\%$ of hidden units are gated on the MNIST dataset for peaked classification accuracy. The optimal gated unit numbers depend on the network complexity and task numbers and need to be tuned by parameter validation. SNN with neuromodulation-assisted credit assignment (NACA) are designed in \cite{zhang2023brain} to mimic the nonlinear long-term synaptic potentiation and depression. NACA employs expectation signals to induce defined levels of neuromodulators to selective synapses of SNN to implement lifelong learning.

However, since the neurons in PFC code for specific tasks and exert top-down control to prioritize context-appropriate stimuli and actions, these methods ignore the dynamic task selectivity of neuron populations. Although some studies explore the context-gating method for lifelong learning \cite{duncker2020organizing,zeng2019continual}, the gated parameters are mostly updated by the designed encoding method or fixation. Hence, we aim to explore the possibility of applying local plasticity to enhance the connection between task-related context input and corresponding activated dendritic synapses, and further integrate global plasticity to achieve lifelong learning of SNN.



\section{Methods}

In this section, we describe the framework of CG-SNN in detail. To explore the applicability of CG-SNN on different SNN' structures and spike-firing mechanisms, we introduce the single-spike and multi-spike CG-SNN here. 
Notably, we analyze and evaluate the performance of the proposed model on the experimental behavior data collected from the relevant cognitive experiments, followed by \cite{flesch2018comparing}. In the above human behavior experimental setting, participants need to make plant or don't plant decisions on the input trees image with continuously varying branch density and leaf density.  The growth success of trees is determined by leaf density in one garden ($1_{st}$ task ) and branch density in the other ($2_{nd}$ task).

\begin{figure*}[!t]
    \centering
\includegraphics[width=2\columnwidth]{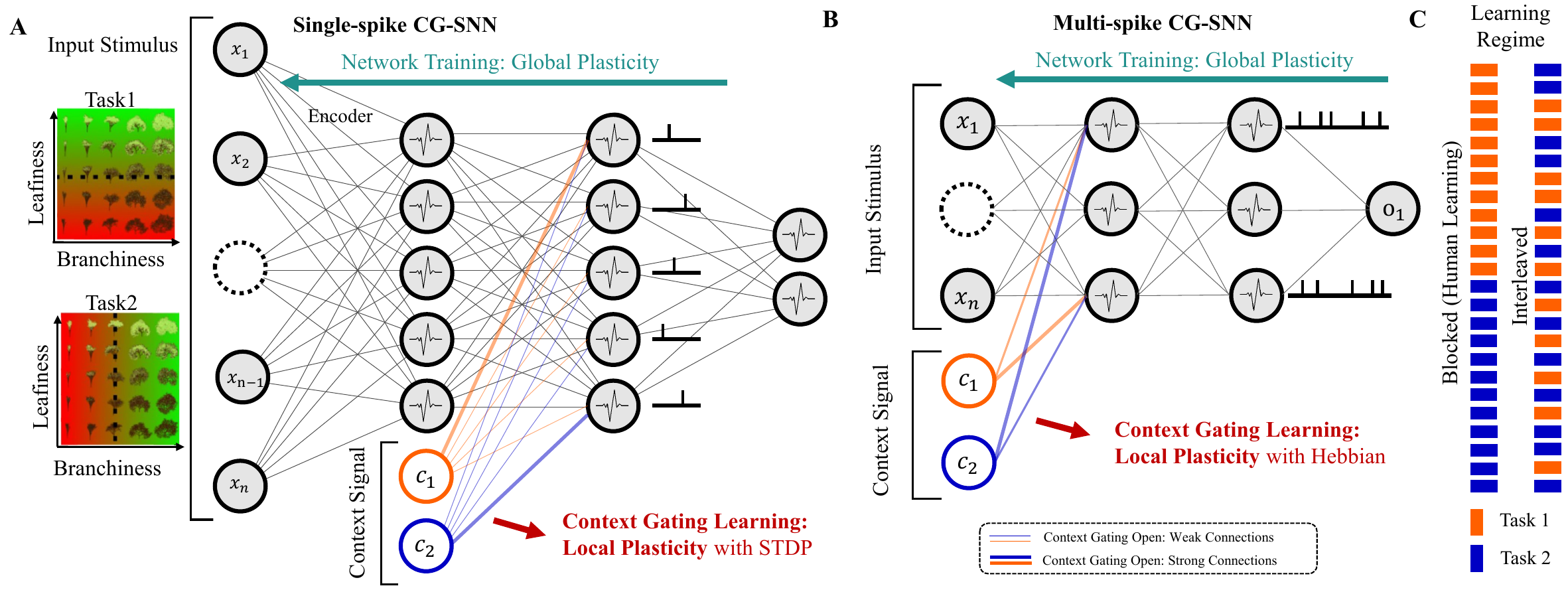}
    \caption{The framework of the network construction of the single-spike CG-SNN and multi-spike CG-SNN. (left) The network structure of single-spike CG-SNN. The first encoding layer codes the input stimuli into spike trains and avoids weak signal ignorance. The additional context signals of tasks are introduced into the second layer. The iterative training between global and local plasticity proceeds during the training process, in which all the synaptic weights are updated by the global backpropagation rule, and the weights of context signal are updated by the local plasticity rule.  
    (middle) The network structure of multi-spike CG-SNN. The network is a simple feed-forward MLP with two hidden layers with IF spiking neurons that receive the pixels information of pictures together with the one-hot contextual signal as input. And the connections between the contextual cue and the first layer will be successively updated by the Hebbian learning rule. (right) shows the blocked and interleaved distribution of the two tasks' samples.}
    \label{fig_model_framework}
\end{figure*}

As shown in Fig. \ref{fig_model_framework}, there are two different kinds of learning regimes: blocked and interleaved regimes. Humans learn better under a blocked curriculum when tasks are not similar, while interleaved training between different tasks could improve the generalization capability of SNN without catastrophic forgetting. 
Thus, we introduce CG-SNN to implement lifelong learning under the block curriculum and model experimental behavior data of humans. 
Through replicating real human behavior data by SNN, the proposed CG-SNN could provide some possible explanation of cognitive control mechanisms from the point view of synaptic plasticity.


\subsection{Network architecture}

We model the above human behavior as three-layer SNN, in which the main part of the network is used to receive tree picture stimuli and output results, while the contextual signals are encoded by one-hot labels and then input into the network to affect the network's output judgment. 

The proposed CG-SNN framework utilizes iterative training to extract and maintain context information for various tasks by alternating between global and local plasticity. The local plasticity is applied to enforce the connection between the context inputs and relevant hidden units. Because the learning of context for each task is essential for multiple-task lifelong learning. Meanwhile, the global plasticity is preserved to update global connection weights, in order to maintain the generalization capability of SNN for a certain task. During the whole training process, dynamic iterative training between global and local plasticity is adopted to balance context learning and generalization learning. Assuming the training epochs of local plasticity and global plasticity are $I_n$ and $I_m$, respectively. We adjust these two parameters by grid search method for single-spike and multi-spike CG-SNN. Through the cross-scale plasticity from local to global, task-related context information could be consolidated by modifying the connection between task neurons and hidden neurons.

Based on different spike-firing methods, we develop CG-SNN based on two different kinds of SNN, multi-spike SNN trained with the surrogate gradient and single-spike SNN with direct backpropagation. Both SNN use stochastic gradient descent (SGD) algorithms to train on blocked or interleaved data. Once a supervised training step is finished, the network will follow a step (some trials more than one) of local plastic learning. 

\subsection{Single-spike CG-SNN} In the network of single-spike CG-SNN, each spike neuron is permitted to fire spike only once.
The non-Leaky IF neuron model is utilized for its simplicity and efficiency \cite{mostafa_supervised_2018}. 
The membrane potential of each postsynaptic neuron $j$ is obtained after integrating the contributions of all the weighted presynaptic neurons of $N_I$:
\begin{equation}
    \label{LIF_V_single_spike}
    V_{j}(t) = \sum_{i=1}^{N_I} g(t-t_i) w_{ij}  (1-exp(-(t-t_i))),
\end{equation}
where $g(x)$ denotes the Heaviside function, which equals zero for negative arguments of $x$, otherwise equals to be one for positive ones. $t_i$ indicates the concrete firing time of presynaptic neuron $i$. Thus, the presynaptic neuron $i$ would transmit the corresponding postsynaptic potential to neuron $j$ only if it satisfies $t_i<t$; otherwise, that potential vanishes. The postsynaptic neuron $j$ emits a spike when its membrane potential crosses the threshold $V_{th}$ which is set to 1. Due to each neuron being permitted to fire only once, the spike firing times $t_j$ is transformed to z-domain by $z_j = exp(t_j) $ to simplify implementation. Meanwhile, the casual set $ C_j = \{ i: t_i < t_j \}$ is defined as the collections of the presynaptic spikes that determine the time point at which the postsynaptic neuron fires the first spike. 
Since then, the firing times of each neuron can be computed sequentially over the feedforward process through the above formulations.

In the experimental process, we find that directly using the single-spike structure in \cite{mostafa_supervised_2018} is easy to make the network ignore the relatively weak signal (that is, the darker pixels) in the source visual stimulus. Therefore, we apply an MLP structure at the first layer, which is used to extract the features of the source image. The hidden layer then encodes the extracted features (the value is limited by implementing a sigmoid function) together with the contextual signal according to their signal strength for forward propagation.

Besides, to strengthen the difference in STDP's influence on different synapses, we try to encode the stronger contextual signal with the same intensity as the strongest signals on the hidden layer during Hebbian updating. We also adjust the number of weight updates of global plasticity and local plasticity in a round in order to coordinate the strength between them.

\textbf{Context gates by STDP.} The local plasticity of STDP is a reasonably well-established physiological mechanism of activity-drive synaptic regulation based on the local adjacent neurons. And it has been observed extensively in vitro for more than a decade \cite{2000Timing}, and it is believed to play an important role in neural activity \cite{2009Competitive}.

\begin{equation}
\Delta W_{ij}^{l} = \begin{cases} 
A_{+} exp (\frac{t_j - t_i}{\tau ^{+}}), \quad   if \ t_j>t_i,\\
A_{-} exp (\frac{t_i - t_j}{\tau ^{-}}),  \quad  if \ t_j<t_i,\\
\end{cases} 
\end{equation} 

where $A_{+}, A_{-}$ and $\tau ^{+}, \tau ^{-}$ indicate the magnitudes and time constants, respectively. Thus, during the local plasticity iteration, the $w_{ij}^{cg}$ is updated by $w_{ij}^{cg} = w_{ij}^{cg} + \lambda_{local} * \Delta W_{ij}^{l} * M_{cg}$. $M_{cg}$ filters the context gating weight connection. During the global plasticity iteration,
since the modeled behavior is a binary classification task, we use a simplified cross-entropy to calculate the loss and then update the global weight connection $W_{ij}^{g}$. Besides, we multiply the loss value by the reward value as the participants would receive the numerical reward after their decision in cognitive experiments.

\begin{equation}
    L_{single} = - r \times \frac{exp(-o_{0})}{exp(-o_{0}) + exp(-o_{1})},
\end{equation}
where $r$ is the corresponding numerical rewards of the trials, $o_{0}$, $o_{1}$ is the spike latency of last layers.



\subsection{Multi-spike CG-SNN} 
%


The global training of the multi-spike CG-SNN model is implemented based on spatiotemporal backpropagation (STBP) algorithm in \cite{wu2018spatio,fang2023parallel,ding2021optimal,zhu2024exploring}. 
The iterative integrate-and-fire (IF) can be employed to express neuronal dynamics both in the spatial domain and temporal domain \cite{fang2023spikingjelly}. According to the integrated presynaptic neuron input and the membrane potential at $t-1$, the membrane potential of postsynaptic neuron $j$ at $t$ can be computed as:

\begin{equation}
    u_i(t+1, n) = u_i(t, n) + x_i(t+1, n) + b_i,
\end{equation}

\begin{equation}
    x_i^{t+1, n} = \sum_{j=1}^{l(n-1)} w_{ij}^{n} o_j^{t+1, n-1},
\end{equation}

\begin{equation}
    o_i^{t+1, n} = g(u_i^{t+1, n} - V_{th}),
\end{equation}
The Heaviside function of $g(x)$ is to guarantee that the neuron emits spike once its membrane potential crosses over the threshold $V_{th}$ (also set to be $1.0$). The leaky IF neuron model also adapts to CG-SNN, where the leaky term with time constant $\tau$ is introduced when computing the above integration term.



Then, the surrogate gradient function of $g_(x)$ is approximated based on the derivative of spike activities of $g'(x) = \frac{\alpha}{2(1 + (\frac{\pi}{2}\alpha x)^2)}$, to solve the non-differential problem of discrete spikes firing behavior, where $\alpha$ is set to be 2.0. Based on the above formulation and the gradient computing in \cite{wu2018spatio, ding2024enhancing}, the update of synaptic weights can be obtained by the gradient descent rules \cite{deng2021comprehensive,zhu2023exploring}.

During experiments, we find that the effect of STDP learning rules in the multi-spike model was very meaningless. We infer that this result might be due to the fact that the positive contextual signal is powerful and is encoded to a relatively high frequency in the multi-spike network. Thus, the local plasticity learned by STDP becomes unstable, which makes STDP rules difficult to assist the network to distinguish functional neurons. Therefore, we don't apply STDP rule to multi-spike model as the local plasticity rule.

\textbf{Oja rules. } Inspired by \cite{flesch2022modelling}, we use Oja's rule \cite{oja1982simplified}, which can regularize the weight based on Hebbian learning, to constrain the parameters:

\begin{equation}
    w_{ij}^{l} = w_{ij}^{l} + \eta_{hebb}y(x - w_{ij}^{l} y),
\end{equation}
where $\eta_{hebb}$ is the learning rate of the Hebbian update. The hyper parameter $\eta_{hebb}$ is obtained through automatic parameter adjustment using Python package of Ray. It is found to be 0.084 when searching from $1e{-4}$ to $1e{-1}$. $x$ is the inputs and $y$ is the linear hidden units output connected to the inputs $x$ via weight matrix $w_{ij}^{l}$. 

\textbf{Sluggish neurons.} According to the ordinary cognition and the conclusion from experiments of human volunteers \cite{flesch2018comparing}, humans can learn tasks better after blocked training than interleaved training, while the artificial neural network is usually on the contrary. Inspired by \cite{flesch2022modelling}, we apply the “sluggish” neurons to model the human's learning bias under interleaved training. That is, when learning in interleaved mode, the relation between the context cues and stimuli of the former samples may interfere with the judgment of the current sample. 
Therefore, sluggish neurons is used to carry information from previous samples over to the current sample, by applying an exponentially moving average to input information of task units. That is, the task-related context signals should updated by $(1-\alpha ) x_t + \alpha x_{t-1}^{EMA}$ when training all other samples ($t>1$) after the first sample ($t=1$).

Through the above description of single-spike and multi-spike CG-SNN, the context gating in the input dimension could imitate the cognitive control mechanism of PFC. The proposed model has higher task-specific neural representation capability because of the extra temporal dynamics and better biological plausibility than artificial neural networks.

\section{Results}

We validate the effectiveness and the biological plausibility of our model on previously published human behavior data in \cite{flesch2018comparing}. The dataset was collected by asking the participant to learn which type of trees would grow well (i.e. give a reward for accept them) in two different gardens. These two gardens with snowy or desert-like environments are two distinct contextual cues/signals. Hence, the input stimuli of CG-SNN are fractal image of trees varied parametrically in trees' branch and leaf density, where only one of these two feature dimensions was relevant in each task and determined grown success (accept reward/penalty). 
CG-SNN performs two tasks' lifelong learning where trees are categoried by leaf and branch density respectively.

\subsection{The effectiveness of single-spike CG-SNN local plasticity with STDP}

\begin{figure*}[!h]
    \centering
    \includegraphics[width=2\columnwidth]{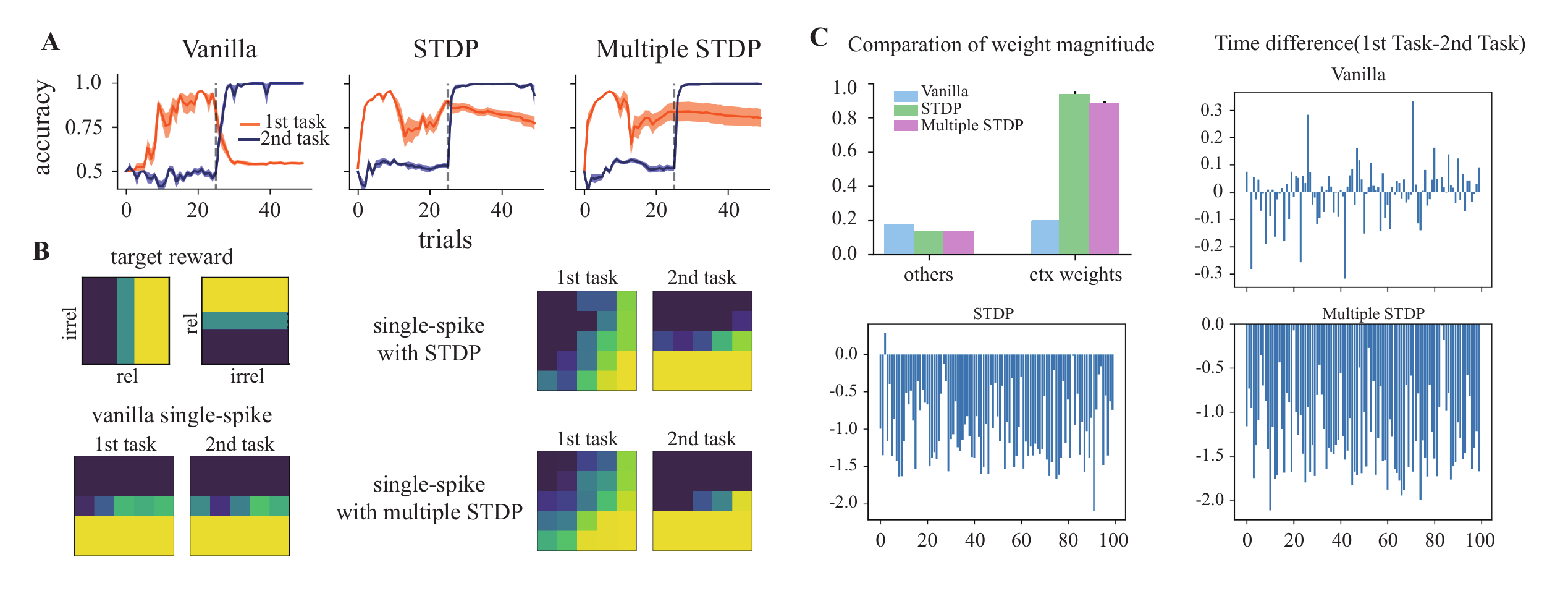}
    \caption{ (A) The accuracy curve of single-spike CG-SNN. (left) Vanilla network of single-spike SNN. (mid) vanilla network with STDP learning rules. (right) vanilla network with multiple STDP update. (B) (left top) The target choice of the given data. (other) Plotting the choice of the trained network with (right top) vanilla setting, (left bottom) STDP learning rules, and (right bottom) multiple STDP updating. (C) (left top) The average absolute values of weights related to the context signal input and other weights in the hidden layer of three networks. (other) The time difference of spiking time of hidden layer neurons of a trained network under two different contextual signals. They are vanilla networks (right top), networks with STDP learning rules (left bottom), and networks with multiple STDP updating (right bottom). }
    \label{combine1}
\end{figure*}

We first explore how STDP may take effect during context-dependent task learning. As illustrated in Fig. \ref{combine1} (A), as expected, we note that single-spike SNN is also affected by catastrophic forgetting, and the learning performance of the first task almost immediately drops down. In contrast, when applying STDP learning rule, the decline rate of the first task is restrained to a very slow level, and the performance of the second task can still rapidly increase until convergence to perfect performance. The accuracy of the first task finally stays at about $\sim$75\%. In addition, the model with multiple STDP updating seems to have stronger resistance to catastrophic forgetting while its results are relatively unstable. We conduct the ANOVA analysis on the slope during the later half epochs to compare multi-stdp and stdp. The p-value is 0.00596 ($<$0.05) and we could conclude that multi-stdp decelerates slower than stdp.

However, although the memory decline decelerates under the effect of STDP, the accuracy curve under the first task becomes even more unstable than the vanilla network. Then we attempt to figure it out by comparing the average absolute values of the weights related to the context signal and other weights in the hidden layer of the three models. As shown in Fig. \ref{combine1} (C) (left top), under the influence of STDP learning rule, the absolute value of context-related weight becomes very large. 
Large weights related to context signal enable the $1_{st}$ task's memory to retain better (these large weights would be more difficult to change much) during training the successive $2_{nd}$ task. 
Besides, after training under the blocked data, the time difference of hidden layer neurons' spike between two tasks is very close (mostly lower than 0.05 ms; the largest is only $\sim$ 0.3 ms) and evenly distributed, which makes the choice matrix of the first task almost replaced by that of the second one. In contrast, the hidden layer neurons of the model using STDP rules always fire spike under the first task earlier than under the second task, which makes the network distinguish the two independent tasks.

Based on these phenomena, we infer that the relatively large weight magnitudes caused by STDP learning rules enable the network to structure some cognition of the past tasks to a certain extent so that it can make correct choices toward relatively strong stimuli after being trained under the second task. 
This can also be reflected in Fig. \ref{combine1} (B) (down), in which the choice matrix is the network's choice (accept reward) made during variety as a function of the two feature dimensions (only a single dimension among branch and leaf is relevant to each task). And the "rel" and "irrel" indicate whether one of these two dimension is related to current task. 
We can see that the choice matrix of the first task smoothly changes along the areas where the choices are consistent under the two tasks. 

Though having a certain effect, the stability and the final performance of the model using STDP learning rules in context-dependent tasks are still not satisfactory. The neural dynamics of the brain are constructed by a large group of neural plasticity  mechanisms \cite{churchland2012neural,ahrens2012brain,golub2018learning}. The exploration and discovery of single-spike CG-SNN might provide some ideas for related research fields. Based on the problems in the single-spike model, we further explore multi-spike CG-SNN, which can implement a more stable and better model of human cognitive control behavior, and has better biological plausibility.

\begin{figure*}[!t]
    \centering
    \includegraphics[width=2\columnwidth]{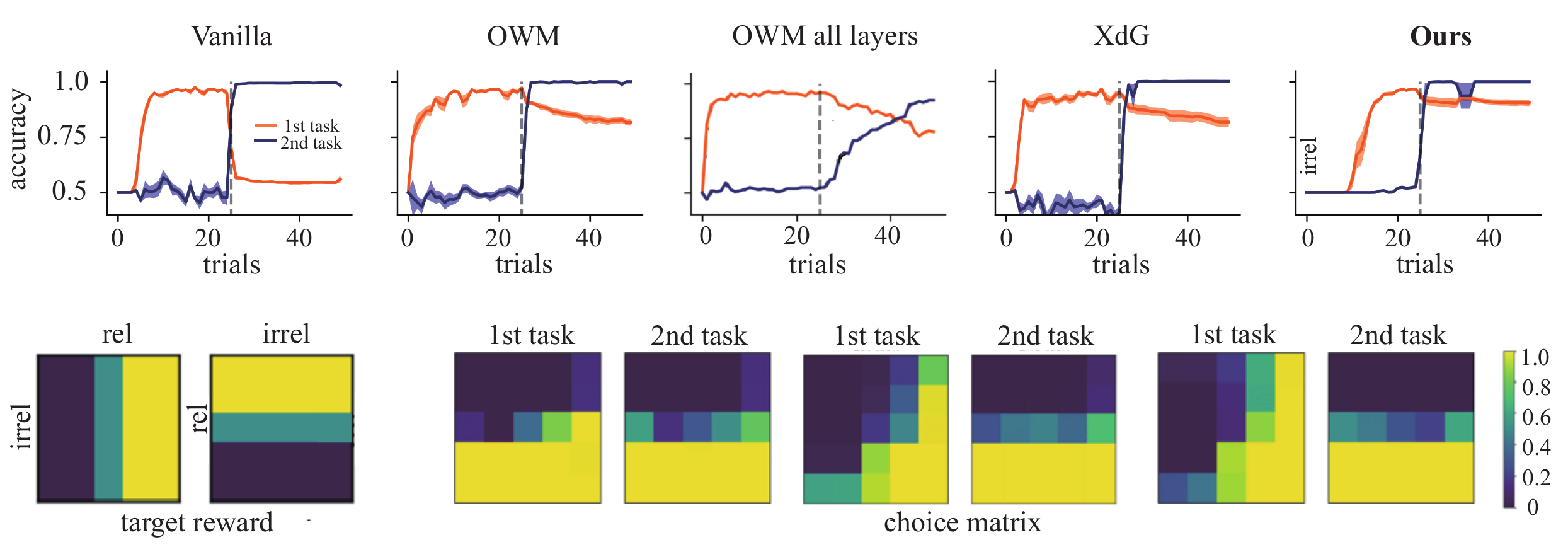}
    \caption{(top) The network accuracy change curve of the vanilla blocked training (left), 
    applying OWM learning algorithm on the latter  two connections and to all the connections (mid
    ),  XdG method, and our method (right). 
    (bottom) The standard reward value of the whole task and the choice of the trained network in two dimensions under the vanilla blocked training network (left two), the network with OWM applied in latter two layers (mid two), and our proposed network (right two).
    }
\label{multi_spike_compare}
\end{figure*}

\subsection{Compared multi-spike CG-SNN with human modification}

To validate the effectiveness of multi-spike CG-SNN, we compare it to other lifelong learning methods under the same setting, such as the regularization-based lifelong learning method of the orthogonal weights modification (OWM) learning algorithm in \cite{zeng2019continual}. As shown in Fig. \ref{multi_spike_compare}, in contrast to the vanilla multi-spike model, the lifelong learning models with the OWM algorithm and our multi-spike CG-SNN perform better on both two tasks, because these models retain context information learning to a certain extent. 
Besides, the performance of the proposed CG-SNN (90.44$\pm$ 3.9\% for task 1, 100\% for task 2) is much better than the XdG (81.69$\pm$ 6.28\% for task 1, 99.94 $\pm$ 0.17\% for task 2), while the XdG model applies more strict constraints on the network connections. It may reveal that excessive human constraint sometimes impacts the network's ability to retain memory. 
In addition, CG-SNN also adapts to SNN with LIF neuron models. The CG-SNN with LIF ($\tau=10.0ms$) achieves 88.69\% $\pm$ 5.19\% and 100\% for task1 and task2, which is also higher than other models.

Moreover, comparing the accuracy curve of the OWM learning algorithm, our multi-spike CG-SNN method can retain the learned tasks to a greater extent under the same circumstances. Especially, the accuracy curve of our method has almost stopped falling, while the curve of the OWM algorithm still shows a downward trend in the last several trials. And on the bottom figure in 
Fig. \ref{multi_spike_compare}
, we can see the detail of the choices made in two dimensions. The vanilla network trained under blocked training almost ignores the task signal, and the OWM model and our CG-SNN model can keep the choices learned from the first task. Meanwhile, the network trained with our method can keep more details near the category boundary of the first tasks.

\subsection{Model the cognitive control behavior of human}


\begin{figure*}[!h]
    \centering
\includegraphics[width=2\columnwidth]{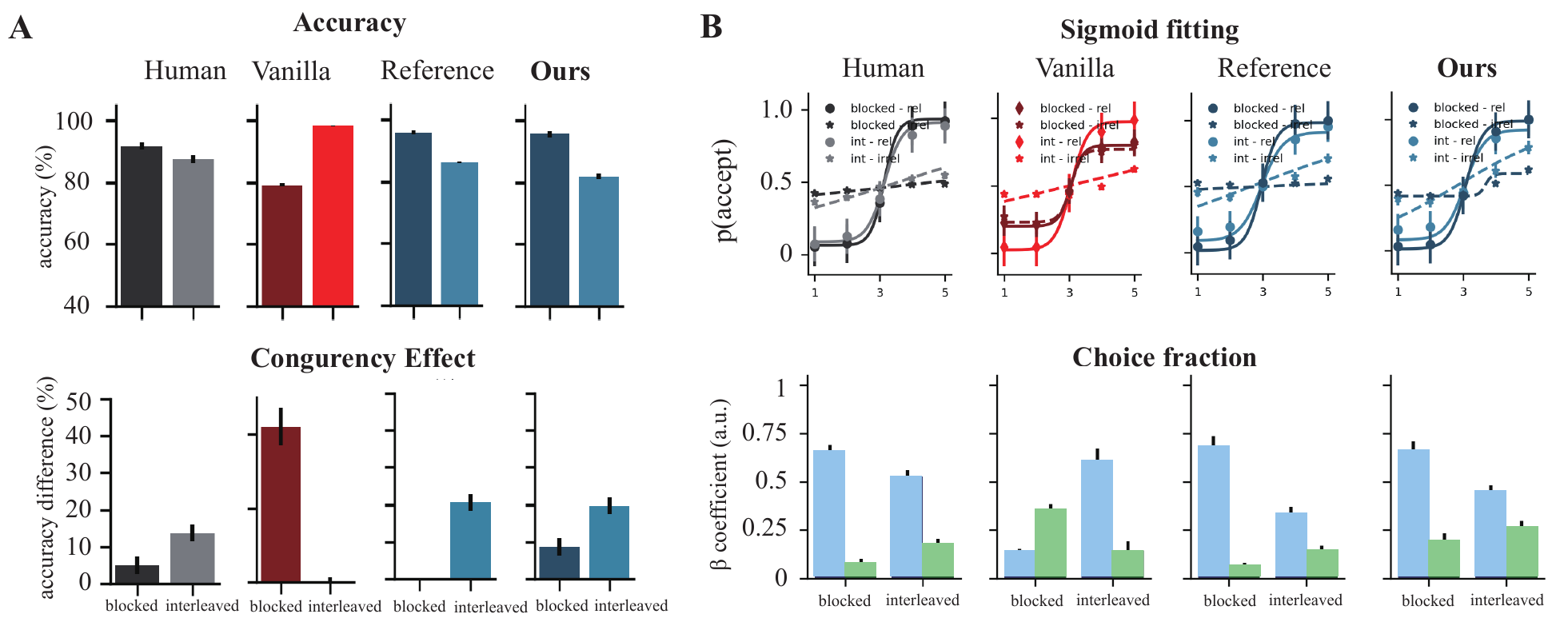}
    \caption{(A) (top) Test accuracy of human data and three network data: (left) human data, (mid left) vanilla network, (mid right) Flesh et al.'s model, (right) our model. (bottom) The comparison of congruency effect. (B)(top) Sigmoidal fits of choice of four conditions mentioned above. (bottom) Linear fitting of factorized and linear model to four conditions. }
    \label{human_compare}
\end{figure*}

We evaluate the effectiveness of our model by comparing with the behavior of the human participants as in \cite{flesch2018comparing}. And to access the rationality and the biological plausibility of our method, we compare the validation performance and neural dynamics after blocked or interleaved training among the experimental data, the vanilla spiking network, the model proposed in \cite{flesch2022modelling}, and our method. 

Firstly, as shown in Fig. \ref{human_compare} (A), we can see that participants trained under blocked data perform better than those trained under interleaved data while the vanilla network shows the opposite phenomenon. The model of \cite{flesch2022modelling} and our model both appear the similarity to human behavior. And participants' behaviors show the congruency effect during the training period, which means that they perform better on the congruent trail between two tasks than the incongruent trials. This effect is stronger when training in the interleaved data, as shown in Fig. \ref{human_compare} (A) (bottom). We can see that the model of \cite{flesch2022modelling} and our model both reveal such a congruency effect while the vanilla network can't reproduce such an effect because of the catastrophic forgetting.

\begin{figure*}[h]
    \centering
    \includegraphics[width=2\columnwidth]{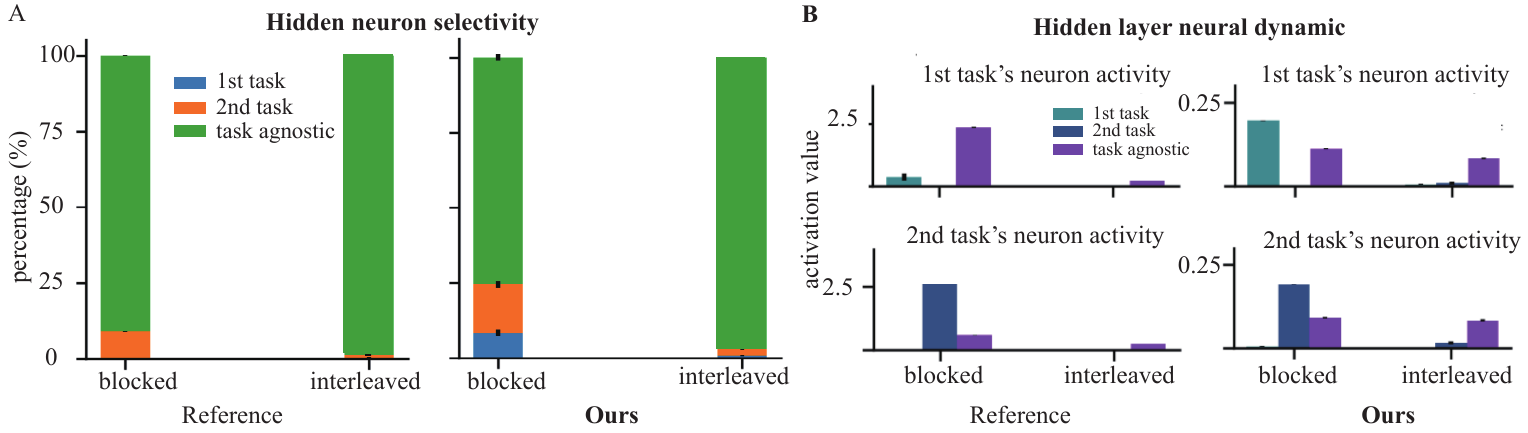}
    \caption{Neuron's selectivity analysis. (left) Proportion of task selective neurons under blocked and interleaved training of the reference network (left) and our CG-SNN networks (right). (right) The average hidden layer neuron's activity of these two networks. The activation values and spiking frequency are computed for reference network and CG-SNN, respectively.}
    \label{fig_neuron_selctivity}
\end{figure*}

Further, we fit the psychometric functions (sigmoid) to compare the choice made by our model and human participants, as shown in Fig. \ref{human_compare} (B) top.
Sigmoid fitting is to fit the sigmoid function to the choice made by humans and CG-SNN models, seperately for the relevant and irrelevant feature dimensions. 
We find that our model and the reference model have relatively steeper slopes for the irrelevant dimension under interleaved training than blocked training as shown in human behavior data. But the vanilla network obtains steeper slopes for the irrelevant dimension under blocked training because the catastrophic forgetting makes the SGD network more easily affected by the irrelevant feature. A similar result also can be seen by fitting the factorised model and the linear model to the choice of human and three networks, as shown in Fig. \ref{human_compare} (B) bottom. 
Choice fraction is obtained by performing the model-based representation similarity analysis on the networks outputs and behavior.
Human participants learned the clear boundary of two tasks/dimensions under blocked data and learned a relatively vague boundary under interleaved data. The reference network and the network with our method can model such characteristics in human behavior while the performance of the vanilla neural network is almost the opposite.
Therefore, our framework can effectively model human cognitive behavior mode and achieve the similar effect to the framework proposed by \cite{flesch2022modelling}. 

\subsection{Analysis the neuron selectivity dynamics}
In addition to modeling human behavior patterns, we also compared the neuron selectivity dynamics in our framework with the ANN model in \cite{flesch2022modelling}, to prove the biological plausibility of our framework. 
As shown in Fig. \ref{fig_neuron_selctivity} (left), 
the ANN model in \cite{flesch2022modelling} shows poor task-selectivity in their network when dealing with trees pictures. 
When under blocked training, their network only has a very small proportion of task-selectivity units in the hidden layer ($\sim$10\%) while in our network a considerable proportion of units is selective to a specific task ($\sim$25\%). 
The number of the neuron units which are only active to the first task even drops to $\sim1\%$. Besides, we plot the average neural activity of task-selective neurons in Fig. \ref{fig_neuron_selctivity} (right).  In our network, the task-selective neurons have a higher spiking frequency than the task-agnostic neurons when facing the corresponding task. The reference network exhibits poor performance on the $1_{st}$ task-selective neuron's activity. 
Since neuron's selectivity reflects the task-specific representation that encodes the relevant dimension of each task under block learning, the better selectivity of CG-SNN implies more powerful resistance to catastrophic forgetting.

\subsection{Model the cognitive control behavior of human}

\begin{figure*}[!h]
    \centering
    \includegraphics[width=2\columnwidth]{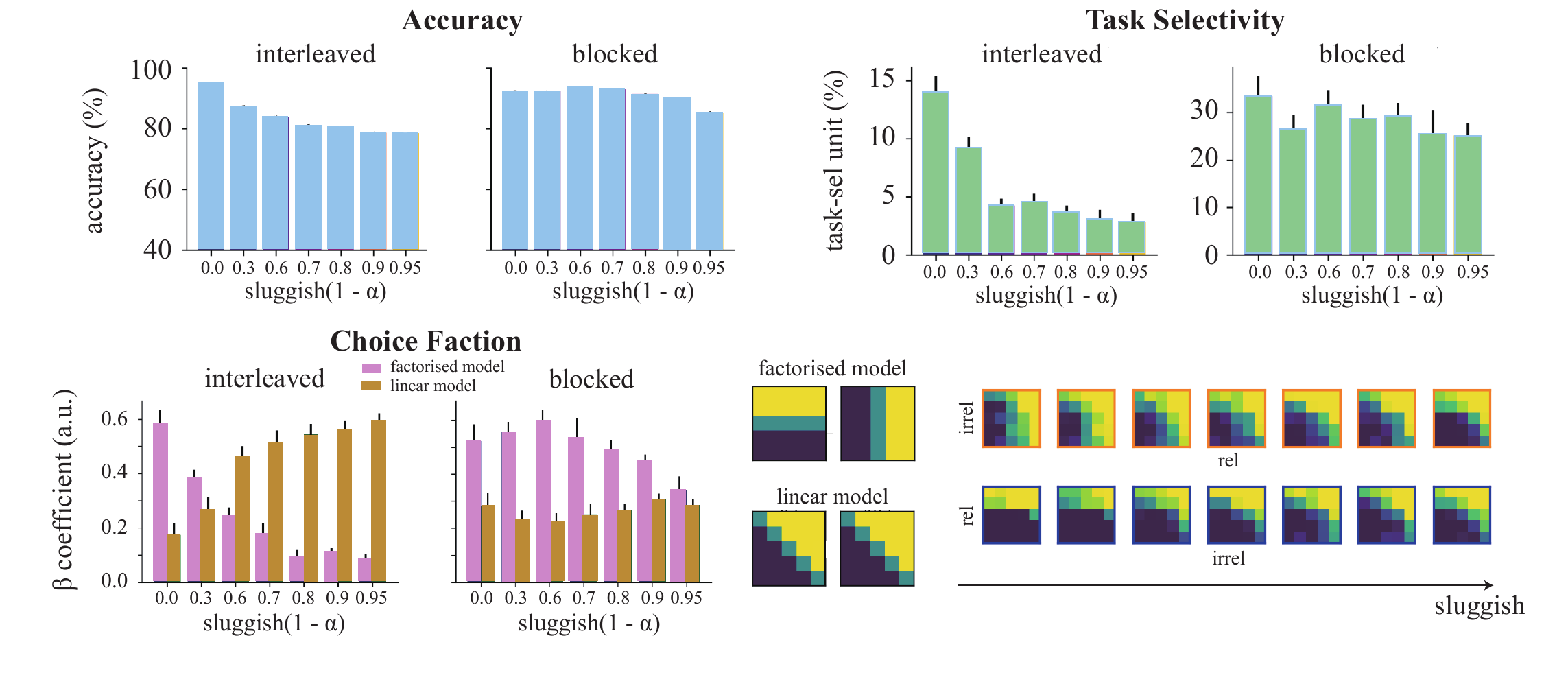}
    \caption{The impact of the sluggish neurons in context-dependent decision making task. (left top) The accuracy of neural networks trained on interleaved data (left) or blocked data (right) with different levels of “sluggishness”. (right top) the proportion of units in the hidden layer which are task selective (obtained by linear regression. The linear regression is Ordinary Least Squares Linear Regression, used to determine task-selectivity of individual neurons). (left bottom) Linear coefficients were obtained from a regression of the output against the models shown in the midden. (mid bottom) The two models are used to model the possible choice of the human. The factorized model means the two features are learned separately while the linear model means the two features are learned chaotically and ignore the context cue. (right bottom) The network output for different levels of sluggishness.
    }
\label{slugish_neuron}
\end{figure*}

We conduct experiments to evaluate the effectiveness of the sluggish neurons. In Fig. \ref{slugish_neuron}, when $\alpha=0$ it means the network is the same as the vanilla network. Combining the results of Fig. \ref{slugish_neuron} (top) and \ref{slugish_neuron} (left bottom), we can conclude that as the level of sluggishness increases the performance of the network decreases.  At the level of neural dynamic, the proportion of the task-selective hidden layer units, which are active only under specific contextual cues, is also reduced. Meanwhile, the choice matrix of the network under interleaved training becomes more and more like the linear model in Fig. \ref{slugish_neuron} (bottom) as the sluggishness values become larger. On the other hand, the same statistical data of the network trained in blocked data with local plasticity hardly be affected by sluggishness. Therefore, the “sluggish” neurons are reasonable assumptions. Meanwhile, though the best accuracy obtained by the trained network under interleaved data (without sluggish) is relatively high, the largest proportion is still lower than that of the trained network under blocked data.


\section{Conclusion}

Affected by catastrophic forgetting, SNN agents often need to learn samples of all tasks in a randomly mixed way to learn new tasks, which is quite different from human multi-task lifelong learning. Therefore, in this paper, we have extensively explored the role of local synaptic plasticity in context gating of SNN, which is more biologically plausible, so as to model human “cognitive control” behavior while minimizing the intervention of artificial inference.
Taken together, based on the gating in the PFC, we studied and explored whether the known local plasticity rules can promote SNN to possess the effect of context gating. The single-spike and multi-spike SNN with context gating trained by local plasticity were implemented. 
The models we proposed have revealed better effectiveness in maintaining the past learning experience. Besides, we fitted our network to the previously published human behavior data and found that our method can reproduce the human behavior data in both blocked and interleaved data. Moreover, our model indicates prior neural dynamics of task neuron selectivity compared with the previous work. 

\bibliographystyle{unsrt}
\bibliography{neurips24}

\begin{thebibliography}{10}

\bibitem{miller2001integrative}
Earl~K Miller, Jonathan~D Cohen, et~al.
\newblock An integrative theory of prefrontal cortex function.
\newblock {\em Annual review of neuroscience}, 24(1):167--202, 2001.

\bibitem{flesch2022modelling}
Timo Flesch, David~G Nagy, Andrew Saxe, and Christopher Summerfield.
\newblock Modelling continual learning in humans with hebbian context gating and exponentially decaying task signals.
\newblock {\em arXiv preprint arXiv:2203.11560}, 2022.

\bibitem{maass1997networks}
Wolfgang Maass.
\newblock Networks of spiking neurons: the third generation of neural network models.
\newblock {\em Neural networks}, 10(9):1659--1671, 1997.

\bibitem{subbulakshmi2021biomimetic}
Shiva Subbulakshmi~Radhakrishnan, Amritanand Sebastian, Aaryan Oberoi, Sarbashis Das, and Saptarshi Das.
\newblock A biomimetic neural encoder for spiking neural network.
\newblock {\em Nature communications}, 12(1):2143, 2021.

\bibitem{kudithipudi2022biological}
Dhireesha Kudithipudi, Mario Aguilar-Simon, Jonathan Babb, Maxim Bazhenov, Douglas Blackiston, Josh Bongard, Andrew~P Brna, Suraj Chakravarthi~Raja, Nick Cheney, Jeff Clune, et~al.
\newblock Biological underpinnings for lifelong learning machines.
\newblock {\em Nature Machine Intelligence}, 4(3):196--210, 2022.

\bibitem{yin2023accurate}
Bojian Yin, Federico Corradi, and Sander~M Bohte.
\newblock Accurate online training of dynamical spiking neural networks through forward propagation through time.
\newblock {\em Nature Machine Intelligence}, pages 1--10, 2023.

\bibitem{pei2019towards}
Jing Pei, Lei Deng, Sen Song, Mingguo Zhao, Youhui Zhang, Shuang Wu, Guanrui Wang, Zhe Zou, Zhenzhi Wu, Wei He, et~al.
\newblock Towards artificial general intelligence with hybrid tianjic chip architecture.
\newblock {\em Nature}, 572(7767):106--111, 2019.

\bibitem{zhang2018highly}
Malu Zhang, Hong Qu, Ammar Belatreche, Yi~Chen, and Zhang Yi.
\newblock A highly effective and robust membrane potential-driven supervised learning method for spiking neurons.
\newblock {\em IEEE transactions on neural networks and learning systems}, 30(1):123--137, 2018.

\bibitem{xu2021robust}
Qi~Xu, Jiangrong Shen, Xuming Ran, Huajin Tang, Gang Pan, and Jian~K Liu.
\newblock Robust transcoding sensory information with neural spikes.
\newblock {\em IEEE Transactions on Neural Networks and Learning Systems}, 33(5):1935--1946, 2021.

\bibitem{fang2021incorporating}
Wei Fang, Zhaofei Yu, Yanqi Chen, Timoth{\'e}e Masquelier, Tiejun Huang, and Yonghong Tian.
\newblock Incorporating learnable membrane time constant to enhance learning of spiking neural networks.
\newblock In {\em Proceedings of the IEEE/CVF international conference on computer vision}, pages 2661--2671, 2021.

\bibitem{fang2021deep}
Wei Fang, Zhaofei Yu, Yanqi Chen, Tiejun Huang, Timoth{\'e}e Masquelier, and Yonghong Tian.
\newblock Deep residual learning in spiking neural networks.
\newblock {\em Advances in Neural Information Processing Systems}, 34:21056--21069, 2021.

\bibitem{mozafari2019bio}
Milad Mozafari, Mohammad Ganjtabesh, Abbas Nowzari-Dalini, Simon~J Thorpe, and Timoth{\'e}e Masquelier.
\newblock Bio-inspired digit recognition using reward-modulated spike-timing-dependent plasticity in deep convolutional networks.
\newblock {\em Pattern recognition}, 94:87--95, 2019.

\bibitem{skatchkovsky2022bayesian}
Nicolas Skatchkovsky, Hyeryung Jang, and Osvaldo Simeone.
\newblock Bayesian continual learning via spiking neural networks.
\newblock {\em arXiv preprint arXiv:2208.13723}, 2022.

\bibitem{allred2020controlled}
Jason~M Allred and Kaushik Roy.
\newblock Controlled forgetting: Targeted stimulation and dopaminergic plasticity modulation for unsupervised lifelong learning in spiking neural networks.
\newblock {\em Frontiers in neuroscience}, 14:7, 2020.

\bibitem{zhang2023brain}
Tielin Zhang, Xiang Cheng, Shuncheng Jia, Chengyu~T Li, Mu-ming Poo, and Bo~Xu.
\newblock A brain-inspired algorithm that mitigates catastrophic forgetting of artificial and spiking neural networks with low computational cost.
\newblock {\em Science Advances}, 9(34):eadi2947, 2023.

\bibitem{hao2023reducing}
Zecheng Hao, Tong Bu, Jianhao Ding, Tiejun Huang, and Zhaofei Yu.
\newblock Reducing ann-snn conversion error through residual membrane potential.
\newblock In {\em Proceedings of the AAAI Conference on Artificial Intelligence}, volume~37, pages 11--21, 2023.

\bibitem{chen2023unified}
Yanqi Chen, Zhengyu Ma, Wei Fang, Xiawu Zheng, Zhaofei Yu, and Yonghong Tian.
\newblock A unified framework for soft threshold pruning.
\newblock {\em arXiv preprint arXiv:2302.13019}, 2023.

\bibitem{hao2023bridging}
Zecheng Hao, Jianhao Ding, Tong Bu, Tiejun Huang, and Zhaofei Yu.
\newblock Bridging the gap between anns and snns by calibrating offset spikes.
\newblock {\em arXiv preprint arXiv:2302.10685}, 2023.

\bibitem{ANTONOV2022512}
D.I. Antonov, K.V. Sviatov, and S.~Sukhov.
\newblock Continuous learning of spiking networks trained with local rules.
\newblock {\em Neural Networks}, 155:512--522, 2022.

\bibitem{grossberg1987competitive}
Stephen Grossberg.
\newblock Competitive learning: From interactive activation to adaptive resonance.
\newblock {\em Cognitive science}, 11(1):23--63, 1987.

\bibitem{masse2018alleviating}
Nicolas~Y Masse, Gregory~D Grant, and David~J Freedman.
\newblock Alleviating catastrophic forgetting using context-dependent gating and synaptic stabilization.
\newblock {\em Proceedings of the National Academy of Sciences}, 115(44):E10467--E10475, 2018.

\bibitem{duncker2020organizing}
Lea Duncker, Laura Driscoll, Krishna~V Shenoy, Maneesh Sahani, and David Sussillo.
\newblock Organizing recurrent network dynamics by task-computation to enable continual learning.
\newblock {\em Advances in neural information processing systems}, 33:14387--14397, 2020.

\bibitem{zeng2019continual}
Guanxiong Zeng, Yang Chen, Bo~Cui, and Shan Yu.
\newblock Continual learning of context-dependent processing in neural networks.
\newblock {\em Nature Machine Intelligence}, 1(8):364--372, 2019.

\bibitem{flesch2018comparing}
Timo Flesch, Jan Balaguer, Ronald Dekker, Hamed Nili, and Christopher Summerfield.
\newblock Comparing continual task learning in minds and machines.
\newblock {\em Proceedings of the National Academy of Sciences}, 115(44):E10313--E10322, 2018.

\bibitem{mostafa_supervised_2018}
H.~Mostafa.
\newblock Supervised {Learning} {Based} on {Temporal} {Coding} in {Spiking} {Neural} {Networks}.
\newblock {\em IEEE Transactions on Neural Networks and Learning Systems}, 29(7):3227--3235, July 2018.

\bibitem{2000Timing}
D.~E. Feldman.
\newblock Timing-based ltp and ltd at vertical inputs to layer ii/iii pyramidal cells in rat barrel cortex.
\newblock {\em Neuron}, 2000.

\bibitem{2009Competitive}
Timothee Masquelier, R.~Guyonneau, and Simon~J. Thorpe.
\newblock Competitive stdp-based spike pattern learning.
\newblock {\em Neural Computation}, 21(5):1259--1276, 2009.

\bibitem{wu2018spatio}
Yujie Wu, Lei Deng, Guoqi Li, Jun Zhu, and Luping Shi.
\newblock Spatio-temporal backpropagation for training high-performance spiking neural networks.
\newblock {\em Frontiers in neuroscience}, 12:331, 2018.

\bibitem{fang2023parallel}
Wei Fang, Zhaofei Yu, Zhaokun Zhou, Ding Chen, Yanqi Chen, Zhengyu Ma, Timoth\'{e}e Masquelier, and Yonghong Tian.
\newblock Parallel spiking neurons with high efficiency and ability to learn long-term dependencies.
\newblock In A.~Oh, T.~Naumann, A.~Globerson, K.~Saenko, M.~Hardt, and S.~Levine, editors, {\em Advances in Neural Information Processing Systems}, volume~36, pages 53674--53687. Curran Associates, Inc., 2023.

\bibitem{ding2021optimal}
Jianhao Ding, Zhaofei Yu, Yonghong Tian, and Tiejun Huang.
\newblock Optimal ann-snn conversion for fast and accurate inference in deep spiking neural networks.
\newblock {\em arXiv preprint arXiv:2105.11654}, 2021.

\bibitem{zhu2024exploring}
Yaoyu Zhu, Wei Fang, Xiaodong Xie, Tiejun Huang, and Zhaofei Yu.
\newblock Exploring loss functions for time-based training strategy in spiking neural networks.
\newblock {\em Advances in Neural Information Processing Systems}, 36, 2024.

\bibitem{fang2023spikingjelly}
Wei Fang, Yanqi Chen, Jianhao Ding, Zhaofei Yu, Timoth{\'e}e Masquelier, Ding Chen, Liwei Huang, Huihui Zhou, Guoqi Li, and Yonghong Tian.
\newblock Spikingjelly: An open-source machine learning infrastructure platform for spike-based intelligence.
\newblock {\em Science Advances}, 9(40):eadi1480, 2023.

\bibitem{ding2024enhancing}
Jianhao Ding, Zhaofei Yu, Tiejun Huang, and Jian~K. Liu.
\newblock Enhancing the robustness of spiking neural networks with stochastic gating mechanisms.
\newblock {\em Proceedings of the AAAI Conference on Artificial Intelligence}, 38(1):492--502, Mar. 2024.

\bibitem{deng2021comprehensive}
Lei Deng, Yujie Wu, Yifan Hu, Ling Liang, Guoqi Li, Xing Hu, Yufei Ding, Peng Li, and Yuan Xie.
\newblock Comprehensive snn compression using admm optimization and activity regularization.
\newblock {\em IEEE transactions on neural networks and learning systems}, 2021.

\bibitem{zhu2023exploring}
Yaoyu Zhu, Wei Fang, Xiaodong Xie, Tiejun Huang, and Zhaofei Yu.
\newblock Exploring loss functions for time-based training strategy in spiking neural networks.
\newblock In {\em Thirty-seventh Conference on Neural Information Processing Systems}, 2023.

\bibitem{oja1982simplified}
Erkki Oja.
\newblock Simplified neuron model as a principal component analyzer.
\newblock {\em Journal of mathematical biology}, 15(3):267--273, 1982.

\bibitem{churchland2012neural}
Mark~M Churchland, John~P Cunningham, Matthew~T Kaufman, Justin~D Foster, Paul Nuyujukian, Stephen~I Ryu, and Krishna~V Shenoy.
\newblock Neural population dynamics during reaching.
\newblock {\em Nature}, 487(7405):51--56, 2012.

\bibitem{ahrens2012brain}
Misha~B Ahrens, Jennifer~M Li, Michael~B Orger, Drew~N Robson, Alexander~F Schier, Florian Engert, and Ruben Portugues.
\newblock Brain-wide neuronal dynamics during motor adaptation in zebrafish.
\newblock {\em Nature}, 485(7399):471--477, 2012.

\bibitem{golub2018learning}
Matthew~D Golub, Patrick~T Sadtler, Emily~R Oby, Kristin~M Quick, Stephen~I Ryu, Elizabeth~C Tyler-Kabara, Aaron~P Batista, Steven~M Chase, and Byron~M Yu.
\newblock Learning by neural reassociation.
\newblock {\em Nature neuroscience}, 21(4):607--616, 2018.

\end{thebibliography}

\vfill

\end{document}